*Article*

# Smart Recommendations for Renting Bikes in Bike Sharing Systems

**Holger Billhardt** [1], **Alberto Fernández** [1,*] **and Sascha Ossowski** [1]

[1] CETINIA, Univ. Rey Juan Carlos, Spain; {holger.billhardt, alberto.fernandez, sascha.ossowski@urjc.es}
* Correspondence: alberto.fernandez@urjc.es

**Abstract:** Vehicle sharing systems, such as bike, car or motorcycle sharing systems, are becoming increasingly popular in big cities in the last years. On one hand, they provide a cheaper and environmental friendlier way of transportation than private cars and, on the other hand, they satisfy the individual mobility demands of citizens better than traditional public transport systems. One of their advantages in this regard is their availability; e.g. the possibility to take (or leave) a vehicle almost anywhere in a city. This availability obviously depends on different strategic and operational management decisions and policies, like the dimension of the fleet or the (re-)distribution of vehicles. Agglutination problems, where due to usage patterns available vehicles concentrate in certain areas whereas no vehicles are available in others, are quite common in such systems and need to be dealt with. Research works have been dedicated to this problem, specifying different techniques to reduce imbalanced situations. In this paper, we present and compare strategies for recommending stations to users who wish to rent or return bikes in station-based bike sharing systems. Our first contribution is a novel recommendation strategy based on queuing theory, that recommends stations based on the utility of the user in terms of lower distance and higher probability of finding a bike or slot. Then we go one step further and define a strategy that recommends stations by combining the utility of a particular user with the utility of the global system, measured in terms of the improvement of the distribution of bikes and slots regarding the expected future demand, with the aim of implicitly avoiding or alleviating balancing problems. We present several experiments to evaluate our proposal with real data from the bike sharing system BiciMAD in Madrid.

## 1. Introduction

Due to the growing need for reducing air pollution and, at the same time, offering mobility services of satisfactory quality to citizens, in many big cities around the world new ways of urban mobility are being explored. Bike sharing systems (BSS) are just one of the many solutions that have been proposed and implemented in recent years. In some cities, they have grown to considerable size. For example, the system Vélib[1] in Paris counts on about 15000 bikes, the Santander Cycles in London[2] has about 11000 bikes and cities in China like Hangzhou or Wuhan even employ 78000 and 90000 bikes, respectively [1]. Some of those systems are so called free-floating systems (e.g. *BikeShare* in Seattle or DB *Callabike* in Munich) that allow citizens to pick up and return a bike at any location within a certain area. Others are station-based, that is, they rely on a set of rental stations with fixed locations. Station-based BSS have the advantage of providing better control to avoid unauthorized access. When electric bikes are deployed in the BSS, as for example in the system BiciMAD in Madrid (Spain), stations usually serve as charging spots for the bicycles' batteries as well.

Effectiveness and user satisfaction within station-based BSS strongly depend on the availability of bikes (or slots for returning rented bikes) at the places where users search them or demand them. Therefore, proper management policies need to be implemented not only for *strategic* decisions (e.g., regarding the positioning and dimensioning of rental stations, the selection of adequate bicycle models, etc.), but also at *operational* level. In particular, BSS are usually faced with agglomeration or unbalancing problems: available bikes are concentrated at certain areas (and at certain times of the day), whereas in other areas only few or no bikes are available. The effect is that users may experience a bad quality of service, being unable to find a bike (or a return slot) at a given station or area. Solutions like reserving bikes (or slots) at some stations palliate this problem, but proper bike balancing mechanisms are needed to attack the problem at its core.

In many systems trucks are used to address such balancing problems by moving bicycles from (almost) full to (almost) empty stations. Truck movements are scheduled either *statically* (usually at night) or *dynamically* (during BSS operation), with the aim of finding a near-optimal match between the expected demand and the available resources. Still, truck movements are expensive and raise environmental concerns.

Nowadays, BSS are often conceived as Cyber-physical Systems [2,3]. Bikes may be endowed with several battery-powered sensors that measure sensible information such as position or battery level (in case of electric bicycles). Sensors at stations may capture rental attempts that were frustrated e.g. due to mechanical failures of slot clamps and, of course, gather key information regarding the operating state of the system, such as the number of available and occupied slots. This information is then transmitted to a software application at the BSS control centre, that helps operators to take timely management decisions in an online manner. Part of this information is often shared with the BSS users who may use smartphone apps that inform them about the current state of stations in real-time. Users can then use this information to take more informed decisions as to where to rent or to return a bike. However, due to the dynamic nature of the BSS environment, it often occurs that, when a user reaches a certain station, initially available bikes have been taken already by other users. Similarly, information on available parking slots may already be outdated when a user arrives to a station. Dynamic reservation schemes can be used to address such problems but may have a negative impact on the overall capacity of a BSS.

BSS are also often perceived from the perspective of the Smart Cities paradigm as part of a more sophisticated heterogenous intelligent transportation solution – a System of Systems that integrates different modes of transport (buses trains, private and shared vehicles, on-demand transportation services, etc.) and enables dynamic collaborations among users such as crowdshipping [4] or crowdsensing [5,6] scenarios. In a multimodal context [7,8], real-time recommendation systems may suggest any type of appropriate transport means (not just bikes) that fit the particular needs and preferences of a user at a particular moment. In some cases, this may alleviate the quality of service problems within a BSS due to imbalances as mentioned above. For example, if there are no available bikes in the area of a user, the system may recommend an on-demand shared taxi service, instead of walking to another station of the BSS. This perspective is very appealing and research in this direction will pave the track towards better and greener transportation services in Smart Cities [9,10]. However, it does not explicitly account for the agglomeration problems that reduce the quality of service at the level of BSS. Therefore, we argue that addressing the imbalance problem in BSS is still a relevant endeavour, as it can impact positively on the overall service level of intelligent transportation systems within a Smart City.

---

[1] http://www.velib-metropole.fr/
[2] tfl.gov.uk/modes/cycling/santander-cycles

In this work we focus on the aforementioned problems within station-based BSS. Our first contribution is a novel recommendation strategy based on a notion of cost that combines *distance* (between a user's origin/destination and a rental/return station) with (an expectation of) *availability* of slots or bikes. To account for the dynamic nature of the BSS environment mentioned earlier, instead of recommending stations that have available bikes or slots at the time a search request is made, we use demand prediction to estimate the expected probability of finding an available bike or slot at the time the user will arrive at a recommended station. In particular, we model stations as queues and apply techniques from queuing theory for probability estimation.

The second contribution of this work is an extension of the aforementioned recommendation strategy that also addresses balancing problems by combining individual objectives with social welfare. For this purpose, the system not only tries to recommend stations with a low cost for the user, but also uses demand predictions to estimate the impact of renting or returning a bike at a given station on society (potential future users). Here, recommendations prioritise stations (for rental or return actions) that combine a low cost for the user with an improved overall impact on the distribution of bikes and slots in the BSS (and, thus, for potential future users).

For evaluating the proposed recommendation strategies, we use a simulation tool for station-based bike sharing systems: Bike3S[3]. This simulator performs semi-realistic simulations of the operation of a bike sharing system in a given area of interest and allows for evaluating and testing different management decisions and strategies. In our simulations we use real data from the BiciMAD BSS which is operating in Madrid (Spain).

The rest of the paper is organised as follows. In Section 2 we discuss related work on bicycle sharing systems. Section 3 outlines the general functioning of a station-based BSS and introduces basic notation used in the rest of the paper. Section 4 focuses on station recommendation strategies that are centred on BSS users. In particular, we sketch some simple straightforward selection strategies before introducing our more sophisticated proposal. Section 5 puts forward several experiments that compare the performance of the different strategies. In Section 6 we present our estimation of the impact of a rental/return action at a station on potential future users, and define our proposal for a station recommendation strategy that combines cost for users and future global impact. In Section 7 we analyse the performance of this strategy in a set of simulation experiments. Finally, Section 8 summarises our work, outlines the lessons we have learnt, and points to future lines of work.

## 2. Related Work

Bike sharing systems management decisions can be taken with different temporal horizons.

Long-term or *strategic* decisions are mainly related to system design and deployment (location and size of stations, number of bikes, etc.) [11-14] as well as the analysis of different exploitation policies for existing systems [15].

At an *operational* level, decisions are taken in order to maintain a good performance of the system, which usually means maximizing the number of users that rent bikes. Thus, actions are aimed at reducing imbalances produced at stations due to heterogeneous demands of bikes and empty slots. To this regard, *off-line* or static approaches are mid-term decisions oriented at repositioning bikes using vehicles (trucks), typically at night or off-peak times, in such a way that bike distribution through stations is optimal for the expected demand. These systems not only estimate the best way of distributing the existing bikes but also the routes taken by the trucks (see e.g., [16-18]).

---

[3] https://github.com/gia-urjc/Bike3S-Simulator

A natural evolution of such systems is their *on-line* application during the actual operation of the fleet. The challenge is combining the current state of the system with demand prediction, so as to propose short-term balancing actions. Many works propose solutions to the problem of dynamically repositioning bikes using trucks as efficiently as possible [19-22]. It has become clear that optimal solutions to the optimization problem do not scale as the size of BSS grows, so authors use heuristics to find solutions applicable in real-world systems.

Furthermore, using trucks to rebalance a bike sharing fleet has several disadvantages such as cost of operation, contribution to congestion and pollution, etc. Recently, several approaches are proposing crowdsourcing this task, so users can contribute by taking or returning bikes at stations different from their target station. Typically, users receive some type of incentive (e.g., discounts) to compensate the extra (usually small) effort of modifying their initial choices.

This paper focuses on dynamic rebalancing systems, so we describe in further detail related work in this area.

Chemla et al. [19] study the problem of obtaining the optimal policy to determine the best action that a repositioning vehicle can take in order to maximise the number of users that find a bike. Since this is an NP-hard problem they propose several heuristic solutions that, roughly speaking, are based on choosing the station with most or least excess of bikes with regard to a given target state. They also employ a pricing strategy in which each station has a price that is paid when the bike is returned, regardless of the origin of the trip. Users choose the destination station based on its current price and the user's walking and biking distances. Their solution consists in optimizing station prices assuming certain user models.

Pfrommer et al. [23] present a dynamic rebalancing system that combines truck-based vehicle redistribution and price incentives for users. In their approach, users either find a bike at their origin stations or leave the system. Upon arrival at their target station users decide whether or not to accept an incentive to return the bike at a neighbouring station. The decision is taken according to their "perceived cost" of the additional distance travelled (biking and walking) and the incentives received. Dynamic prices are calculated through an optimization problem that combines the quality of the resulting state and the cost of paying out the incentives. The key component of the system quality is the analysis of each station's utility based on an optimal "fill level" (the available bikes at a station).

Haider et al. [24] propose a method that combines trucks and price incentives. In their approach, they intentionally try to make some stations even more imbalanced (hubs), so bike redistribution can be carried out more efficiently with short time truck trips. The objective is to minimize the number of imbalanced stations, which they solve heuristically. Prices (which include incentives) are established for O-D station pairs.

Singla et al. [25] present a mechanism that incentivizes the users in the bike repositioning process. They employ optimal pricing policies using regret minimization in online learning, and investigate the incentive compatibility of their mechanism. The approach is evaluated through simulations based on historical data from the Boston Hubway system as well as on data collected via a survey study with MVGmeinRad in Mainz, Germany.

Reiss and Bogenberger [26] propose a hybrid approach for a free-floating bike sharing system in which user's discounts (100-60%) are applied when the imbalance is low (less than 15%) or trucks are used in case of high imbalances. The decision is updated once for each of five predefined time slots during the day. Imbalances are measured as the difference between the current and a desired fleet distribution. The latter is calculated for the entire day from historical data.

Wang and Hou [27] propose a repositioning strategy that assumes there is a number of voluntary riders at each time period. Trips (including bike repositioning by volunteers) are considered for time periods of 30 minutes. The authors use integer programming to solve two optimization problems: first (i) the "ideal bike inventory" for each station and time period (with unlimited voluntary riders), then (ii) the voluntary rider flow (OD pairs) so as to obtain an inventory as close to the "ideal" as possible.

Chung et al. [28] analyse Bike Angels, the incentive program of New York City's Citi Bike system. Angels are users that earn points and rewards for helping rebalancing the system. The authors analyse the Bike Angels program and propose several offline and online incentivizing policies, one of which has been adopted by Citi Bike.

O'Mahony [29] designs a raffle-based incentive system in which users can gain tickets for taking bikes from surplus stations and returning them at shortage stations. According to [30], this type of incentives is more effective than micro incentives (e.g., few cents discounts).

Fricker and Gast [31] study homogeneous BSS and calculate the optimal fleet size to minimise the proportion of problematic stations (empty or full). They conclude that simply returning bikes to non-saturated stations does not produce a sufficient impact on the system behaviour. However, the mere incentive of suggesting users to return bikes at the "worst" among (only) two stations (chosen at random) improves the system dramatically, even if only some of the users follow the recommendations.

Chiariotti et al. [32] propose a solution that combines rebalancing (with trucks) and incentivising users with rewards and prizes, as this combination has a lower cost than using only trucks. Using a finite Markov birth-death process, their approach is to minimise the average expected time that stations are empty or full. They are only interested in the overall effect of incentives on the system rather than of specific types of incentives.

Li and Shan [33] propose a bidirectional incentive model where two types of users are considered, namely commuters and leisure travellers. The users' travel behaviours are characterised as *peak*, *flat peak* or *inverse peak* travel. The goal of the system is to persuade commuters to move from *peak* to *flat peak* behaviour, and leisure travellers from *flat peak* to *inverse peak*. This change in behaviour is incentivised by charging more to commuters or rewarding leisure users, respectively. Subsidies from the government are also included in their strategies.

User behaviour models are also relevant to static rebalancing approaches [22]. In that work, the behaviour of users when arriving at empty stations is simulated, i.e., if they leave the system or walk to another station. Based on a survey the authors calculate the rate of users that are willing to walk to a neighbouring station depending on the existing distance.

In the present work we do not focus on explicitly incentivizing users but assume that they follow the recommendations made. As users gain experience with our recommendation service, we expect them to learn that it helps them achieve a better user experience and to establish trust that the suggestions made are aligned with their goals. To this respect, our work is similar to [34], who created Cityride, a bike sharing journey advisor for the city of Dublin. In their system, origin and destination stations are recommended based on the overall travel time and the probability to find a bike at origin and a slot at destination stations at the respective expected arrival times. Even though it is not the primary goal of recommendations to rebalance the system, this is achieved indirectly as a side-effect.

With respect to the applied mathematical tools and methods, our approach is related to the work by Waserhole and Jost [35], who theoretically study the regulation of station-based vehicle sharing systems through pricing using queuing networks. However, their study makes a series of assumptions that hinder its applicability, namely null transportation times, infinite capacities of stations and stationary demand over time.

As in the present paper, López Santiago et al. [36] also use historical data from the BiciMAD BSS in Madrid. They use social simulations to analyse the impact of several simple price incentive schemes and compare them to the policy currently employed in the BiciMAD BSS. Their work mainly focuses on adequately modelling expected user behaviour and, in particular. the willingness to participate in rebalancing for given distances and incentives, based on existing surveys. In contrast, in the present work we aim at systematically determining efficient solutions based on queuing theory as the basis for recommendations.

### 3. Station-Based BSS

A station-based BSS consists of a set of stations, distributed around the city at different, fixed geographical locations. Each station has a number of slots to park a bike (its capacity), which may differ from station to station. At a given point in time, some slots may be empty or available (e.g., can be used to return a bike) and others have an available bike plugged in that can be taken by a user. The total sum of bikes in the system is always much smaller than the sum of capacities of all stations, so as to allow users to find empty slots to return bikes.

Users who want to use the system would go to a station, take a bike and return it after some time at another station close to the location of her destination. Normally, when a user decides to take a bike, she would be interested in finding the one closest to her initial position. In the same way, when she wants to return a rented bike, she would like to do that at an available slot as close as possible to her final destination.

Usually, BSS are supported by software applications that provide real-time information about the current state of stations to the users. Users may consult them in order to find appropriate stations to rent or to return a bike. In some cases, recommendation systems exist that propose appropriate stations to users. However, even if a user applies this real-time information to determine appropriate stations with available bikes or slots, it may occur that when she actually arrives at that station, the initially available bikes (or slots) have already been taken (or occupied) by other users. We call this situation an *unsuccessful* rental (or return) attempt. If a user has arrived at a station and there are no bikes available, she might consider going to a neighbouring station and try there, or she may desist. A user will usually desist if she does not find an available bike after walking a certain distance (and perhaps trying at several stations). She might also desist if, based on the real-time information provided by her software application, she realizes that there are currently no bikes available within a reasonable distance to her initial position. Once a user has rented a bike, she is obliged to return it at some station. That is, there is no possibility to desist: the user has to go to another station if there are no available slots at the station where she has tried first.

In this work, we intend to provide users with a recommendation system that suggests the "best" station to them when they want to rent a bike in the system (or they have to return a rented bike). That is, we assume that potential users, when they decide to rent a bike or to return a rented bike, consult our system for recommendations of "good" stations. Furthermore, users would usually follow the given recommendation and move to a recommended station. In this sense, our aim in this paper is to define and analyse recommendation strategies that, if followed by the users, provide improved performance at individual and at social (system) level. The general objective is to recommend stations that i) are likely to have the requested resource (bike or slot) once a user arrives there, and ii) provide short travel times, e.g., are close to the location of the user in case of a bike rental or close to her final destination in case of a bike return.

In the rest of the paper, we use the following notations (see Table 1): $S = \{s_1, \ldots, s_n\}$ denotes the set of stations, each with geographical location $l(s_i)$ and capacity $c(s_i)$. $bikes(s_i, t)$ and $slots(s_i, t)$ represent the number of available bikes and slots at station $s_i$ at time $t$, respectively. A user $u_k$ who wishes to rent out a bike will request a recommendation for an appropriate rental station. Such a request, denoted by $rent_k = \langle l_k, t, md \rangle$ consists of the current (or desired origin) location $l_k$ of the user, the current time $t$ when the request is issued, and the maximum distance $md$ a user would be willing to walk to get a bike. If there is no station within distance $md$, the user would desist and abandon the system. In a similar way, a user who wishes to return a rented bike will request a recommendation for a station to return that bike. This request is denoted by $return_k = \langle l_k, d_k, t \rangle$ and consists of the current location $l_k$, the issuing time $t$, and the location of the user's final destination $d_k$. In the case of a return request, a maximum distance does not apply, since the user is required to return the bike in any case. $wtime(x, y)$ and $btime(x, y)$ denote the expected time and $wdist(x, y)$ and $bdist(x, y)$ the expected distance to walk/cycle from location $x$ to location $y$. Both time and distance measures depend on the walking and cycling velocities of each user, and we assume that they can be estimated using standard velocities.

We define rental and return recommendations as functions that map a rental or return request ($rent_k = \langle l_k, t, md \rangle$ or $return_k = \langle l_k, d_k, t \rangle$) to a ranking of stations (e.g., an ordered list with pairwise distinct elements): $RentStation(rent_k) = (s_1, \ldots, s_m) = (s_i)_{i=1}^{m}$ and $ReturnStation(return_k) = (s_1, \ldots, s_m) = (s_i)_{i=1}^{m}$ with $s_i \in S$.

Table 1. List of symbols and functions used.

| Symbols | Description |
|---|---|
| $S = \{s_1, \ldots, s_n\}$ | Set of stations |
| $l(s_i)$ | Location of station $s_i$ |
| $c(s_i)$ | Capacity of station $s_i$ |
| $bikes(s_i, t)$ | Number of available bikes at time $t$ |
| $slots(s_i, t)$ | Number of empty slots at station $s_i$ at time $t$ |
| $u_k$ | User $k$ |
| $rent_k = \langle l_k, t, md \rangle$ | Rental request of user $k$, with: $l_k$ (initial user location), $t$ (time the request is issued), $md$ (maximum distance the user is willing to walk to get a bike) |
| $return_k = \langle l_k, d_k, t \rangle$ | Return request of user $k$, with: $l_k$ (initial user location), $d_k$ (location of the user's final destination), $t$ (time the request is issued) |
| $wtime(x, y)$ | Expected walking time from $x$ to $y$ |
| $btime(x, y)$ | Expected cycling time ¡ from $x$ to $y$ |
| $wdist(x, y)$ | Expected walking distance from $x$ to $y$ |
| $bdist(x, y)$ | Expected cycling distance from $x$ to $y$ |
| $RentStation(rent_k)$ | Ordered sequence of recommended rental stations $(s_1, \ldots, s_m) = (s_i)_{i=1}^{m}$ |
| $ReturnStation(return_k)$ | Ordered sequence of recommended return stations $(s_1, \ldots, s_m) = (s_i)_{i=1}^{m}$ |
| $t_{exp}$ | Expected arrival time of a user at a station |
| $t_x$ | Time of the last known expected event |
| $changes(s_i, t, t_{exp})$ | Expected change in number of bikes in station $s_i$ in interval $[t, t_{exp}]$ with respect to the number of bikes at time $t$ (e.g. +2 indicates there will be two more bikes at time $t_{exp}$) |
| $committedSlots(s_i, t)$ | maximum number of "committed" slots at station $s_i$ after time $t$. This value is $\geq 0$ |
| $committedBikes(s_i, t)$ | maximum number of "committed" bikes at station $s_i$ after time $t$. This value is $\geq 0$ |
| $estimatedBikes(s_i, rent_k)$ | Number of expected available bikes at $s_i$ at the time user $k$ is expected to arrive and taking into account the expected changes and "committed" bikes |
| $estimatedSlots(s_i, return_k)$ | Number of expected available slots at $s_i$ at the time user $k$ is expected to arrive and taking into account the expected changes and "committed" slots |

## 4. User-Centred Station Recommendation

In this section we define several strategies for recommending the best stations for renting/returning a bike from a user's perspective. We first present simple straightforward versions of such user-centred recommendation strategies as baselines, and then develop a smarter strategy that forecasts expected changes in order to improve the estimation of available bikes/slots for the time when a user will actually arrive at a station. Finally, we present an even more sophisticated strategy that estimates the probability of finding available bikes/slots based on current demand data and uses this information to calculate a "cost" for renting or returning a bike at a station.

*4.1. Standard Strategies*

4.1.1. Shortest Distance

This is a basic selection strategy, where a user simply prefers the station that is closest to her origin/destination position. This strategy represents the behaviour of a user who just goes to the closest station, without knowing whether there are available bikes or slots.

$$RentStation(rent_k) = (s_i)_{i=1}^m, \text{ where} \tag{1}$$
$$\forall s_i: wdist(l_k, l(s_i)) \leq md \land \forall s_i, s_j, i < j: \ wdist(l_k, l(s_i)) \leq wdist(l_k, l(s_j))$$

$$ReturnStation(return_k) = (s_i)_{i=1}^m, \text{ where} \tag{2}$$
$$\forall s_i, s_j, i < j: \ wdist(l(s_i), d_k) \leq wdist(l(s_j), d_k)$$

4.1.2 Informed Shortest Distance

In this strategy, the system proposes the closest stations that have available bikes (or slots). Thus, a user avoids going to stations that are empty (at the moment of station selection). However, a station may no longer hold available bikes or slots at the time when the user actually arrives to it, because other users have taken the available bikes (or have occupied the available slots).

$$RentStation(rent_k) = (s_i)_{i=1}^m, \text{ where} \tag{3}$$
$$\forall s_i: wdist(l_k, l(s_i)) \leq md \land bikes(s_i, t) > 0 \land$$
$$\forall s_i, s_j, i < j: \ wdist(l_k, l(s_i)) \leq wdist(l_k, l(s_j))$$

$$ReturnStation(return_k) = (s_i)_{i=1}^m, \text{ where} \tag{4}$$
$$\forall s_i: slots(s_i, t) > 0 \land \forall s_i, s_j, i < j: \ wdist(l(s_i), d_k) \leq wdist(l(s_i), d_k)$$

4.1.3 Distance Resources

Here, stations are selected by combining distance and available resources. The preferred stations are the ones that are closer to the origin (or destination) location of the user and have more available bikes (slots) at the time a recommendation is requested.

$$RentStation(rent_k) = (s_i)_{i=1}^m, \text{ where} \tag{5}$$
$$\forall s_i: wdist(l_k, l(s_i)) \leq md \land bikes(s_i, t) > 0 \land$$
$$\forall s_i, s_j, i < j: \ \frac{wdist(l_k, l(s_i))}{bikes(s_i, t)} \leq \frac{wdist(l_k, l(s_j))}{bikes(s_j, t)}$$

$$ReturnStation(return_k) = (s_i)_{i=1}^m, \text{ where} \tag{6}$$
$$\forall s_i: slots(s_i, t) > 0 \land \forall s_i, s_j, i < j: \ \frac{wdist(l(s_i), d_k)}{slots(s_i, t)} \leq \frac{wdist(l(s_j), d_k)}{slots(s_j, t)}$$

*4.2. Distance Expected Resources*

If a recommendation or information system is used by users to select a station to rent or return a bike, this system could take into account past requests in order to make better estimations of available bikes at a station in the near future. In particular, if a user $u_k$ asks at time $t$ to rent or return a bike, the system may recommend a station $s_i$ and can assume that the user will take or return a bike there when she arrives, i.e. at time $t + wtime(l_k, l(s_i))$ or $t + btime(l_k, l(s_i))$, respectively. This information can be used to better determine whether or not there will be available bikes at $s_i$ if another user requests information. In the same way, expected returns of bikes can help to predict available slots.

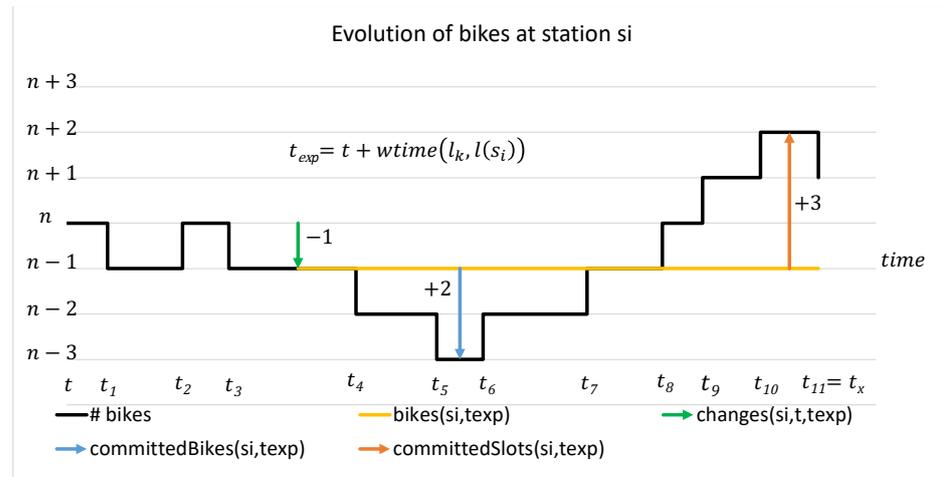

**Figure 1.** Example of number of bikes evolution at a station. Values of $changes(s_i, t, t_{exp})$, $commitedBikes(s_i, t_{exp})$ and $commitedSlots(s_i, t_{exp})$ are shown.

Figure 1 explains this idea in more detail. Suppose that at time $t$ user $u_k$ (at location $l_k$) asks for a station to rent a bike and that one option would be station $s_i$. At time $t$, there are $n$ available bikes at station $s_i$ and the system tries to estimate the available bikes at the moment when $u_k$ would actually arrive at $s_i$, i.e. at time $t_{exp} = t + wtime(l_k, l(s_i))$. Before instant $t$, the system has already recommended station $s_i$ to other users for both, renting or returning bikes, and assumes that those users will actually follow the recommendations and thus, will return or take bikes at their expected arrival times. These actions are reflected in Figure 1: at time $t_1$ a bike is expected to be taken, at $t_2$ a bike is expected to be returned and so on. Let $changes(s_i, t, t_{exp})$ denote the sum of the expected changes of available bikes at station $s_i$ in the time interval between the time of issuing a user request $t$ and the potential arrival of the user at $t_{exp}$. Here each bike rental in this period, counts –1 and each bike return counts +1. In the example, $changes(s_i, t, t_{exp})$ would be –1. We can now estimate the number of bikes or slots expected to be available at the arrival time $t_{exp}$ of user $u_k$ by $bikes(s_i, t) + changes(s_i, t, t_{exp})$ and $slots(s_i, t) - changes(s_i, t, t_{exp})$.

The system may also take into account that even if there are some bikes or slots available when user $u_k$ arrives, some of those may have been already "committed", e.g., they have been the basis for recommendations to other users who will eventually arrive after the expected arrival time of user $u_k$. In Figure 1, this refers to the events at $t_4, t_5, \ldots t_x$. Such events could be conceived as "implicit reservations" and the system could try to "block" a number of bikes or slots for further recommendations. Here we do not refer to a real blocking of a bike or slot for future users. Instead, the system would just include this information when determining recommendations.

The number of "committed" bikes or slots depends on the sequence of expected events and corresponds to the worst cases: the maximum and minimum values of $changes(s_i, t_{exp}, t_x)$, the expected changes after the user arrives and up to the last known event at time $t_x$ (or $t_x = t_{exp}$, if there is none). We denote these two values by

$$committedSlots(s_i, t_{exp}) = \max_{t_{exp} \leq t_i \leq t_x} \left( changes(s_i, t_{exp}, t_i) \right) \quad (7)$$

and

$$committedBikes(s_i, t_{exp}) = \left| \min_{t_{exp} \leq t_i \leq t_x} \left( changes(s_i, t_{exp}, t_i) \right) \right| \quad (8)$$

In Figure 1 $commitedSlots(s_i, t_{exp}) = 3$ denotes that three extra slots have to be "reserved" because a sequence of two taken and five returned bikes reaches a peak of 3 before $t_x$. On the other hand, $commitedBikes(s_i, t_{exp}) = 2$ indicates that at some point (in this case at $t_5$) two extra bikes will be taken from the station.

Taking into account the changes expected to occur at a station $s_i$, after $t$ and before the expected arrival of user $u_k$, as well as the "committed" bikes and slots, the estimated available bikes and slots at station $s_i$ for a rental and a return request $rent_k = \langle l_k, t, md \rangle$ and $return_k = \langle l_k, d_k, t \rangle$ are defined as:

$$estimatedBikes(s_i, rent_k) = bikes(s_i, t) + changes(s_i, t, t_{exp}) - commitedBikes(s_i, t_{exp}) \quad (9)$$

$$estimatedSlots(s_i, return_k) = slots(s_i, t) - changes(s_i, t, t_{exp}) - commitedSlots(s_i, t_{exp}) \quad (10)$$

where $t_{exp}$ is the estimated expected arrival time of user $u_k$ at $s_i$.

With these definitions, we can define the *DistanceExpectedResources* strategy as follows:

$$RentStation(rent_k) = (s_i)_{i=1}^m, \text{ where} \quad (11)$$
$$\forall s_i: wdist(l_k, l(s_i)) \leq md \wedge estimatedBikes(s_i, rent_k) > 0 \wedge$$
$$\forall s_i, s_j, i < j: \frac{wdist(l_k, l(s_i))}{estimatedBikes(s_i, rent_k)} \leq \frac{wdist(l_k, l(s_j))}{estimatedBikes(s_j, rent_k)}$$

$$ReturnStation(return_k) = (s_i)_{i=1}^m, \text{ where} \quad (12)$$
$$\forall s_i: estimatedSlots(s_i, return_k) > 0 \wedge$$
$$\forall s_i, s_j, i < j: \frac{wdist(l(s_i), d_k)}{estimatedSlots(s_i, return_k)} \leq \frac{wdist(l(s_j), d_k)}{estimatedSlots(s_j, return_k)}$$

### 4.3. Expected Cost

It seems obvious that a better estimation of the available bikes or slots at the time of arrival of a user at a station would avoid unsuccessful rental or return attempts. One way to achieve this is by estimating the probability distributions of available bikes at a station at a future time instant, based on known demand patterns. In the following we describe how demands can be estimated, how we estimate the probabilities of finding an available bike/slot at a station at any time in the future, and how we use such probabilities to rank stations increasingly based on cost, combining the probability of finding a bike/slot and the expected time to reach a station.

Table 2 summarises the symbols and functions introduced in this section.

**Table 2.** List of symbols and functions used.

| Symbols | Description |
|---|---|
| $rentDemand(s_i, t_a, t_b)$ | Number of expected rent attempts at $s_i$ between $t_a$ and $t_b$ |
| $returnDemand(s_i, t_a, t_b)$ | Number of expected return attempts at $s_i$ between $t_a$ and $t_b$ |
| $\pi^i(t)$ | (Transient) State probability distribution of the number of bikes at station $s_i$ at time $t$ |
| $\pi_j^i(t)$ | Probability that there are $j$ available bikes at station $s_i$ at time $t$ |
| $bikeProb(s_i, rent_k)$ | Probability that $u_k$ would find an available bike at station $s_i$ when she arrives |
| $slotProb(s_i, return_k)$ | Probability that $u_k$ would find an available slot at station $s_i$ when she arrives |
| $\lambda$ | Average bike arrival rate |
| $\mu$ | Average bike rental rate |
| $RentFailCost$ | cost of getting to a station and finding no available bike |
| $ReturnFailCost$ | cost of getting to a station and finding no available slot |
| $localRentCost(s_i, rent_k, RentFailCost)$ | Estimated cost of user $u_k$ to rent a bike at $s_i$, resulting of combining walking time to the station and the probability of finding an available bike |
| $localReturnCost(s_i, return_k, ReturnFailCost)$ | Estimated cost of user $u_k$ to return a bike at $s_i$, resulting of combining cycling time to the station, walking to final destination and the probability of finding an available slot |

4.3.1. Demand Estimation

The demand of bike rentals and returns at a station can be estimated from historical data and for different time intervals, days, weather conditions, and so on. Supposing that such data is available, we use $rentDemand(s_i, t_a, t_b)$ and $returnDemand(s_i, t_a, t_b)$ to reflect the number of expected rental and return attempts at station $s_i$ during the time interval between $t_a$ and $t_b$ ($t_a < t_b$). Given sufficient historical data, the calculation of these values is straightforward and we omit it here.

4.3.2. Probability calculation

Using queueing theory, a station for renting bikes can be modelled as a M/M/1/K queue (see, e.g., [21,31,35]). We consider that bikes arrive at a station and request the "service" to be rented. The arrival process (e.g., the return of bikes through users) and also the "service" process (e.g., the rental of bikes by users), follow a Poisson distribution. The number of service channels is 1 and the maximum number of bikes waiting to be rented is K (the capacity of the station). The M/M/1/K queue is a special case of a birth-death process and of continuous time Markov chains with K+1 states. Let $\lambda$ be the average bike arrival rate and $\mu$ the average bike rental rate. Given a station $s_i$, let $\pi^i(t)$ denote the (transient) state probability distribution and $\pi_j^i(t)$ the probability that the system is in state $j$ at time $t$. That is, $\pi_j^i(t)$ represents the probability that there are $j$ available bikes at station $s_i$ at time $t$.

The time-dependent behaviour of the probability distribution can be determined by the system of differential equations (known as the forward Kolmogorov equations):

$$\frac{d\pi_0^i(t)}{dt} = -\lambda \pi_0^i(t) + \mu \pi_1^i(t) \tag{13}$$

$$\frac{d\pi_j^i(t)}{dt} = \lambda \pi_{j-1}^i(t) + \mu \pi_{j+1}^i(t) - (\lambda + \mu)\pi_j^i(t), \quad \text{for} \quad 0 < j < K \tag{14}$$

$$\frac{d\pi_K^i(t)}{dt} = \lambda \pi_{K-1}^i(t) - \mu \pi_K^i(t) \tag{15}$$

Given an initial distribution $\pi^i(0)$ with $\sum_{j=0}^{K} \pi_j^i(0) = 1$ the system can be solved obtaining the distribution $\pi^i(t)$ at time $t$. Solving (1) is mathematically complex, but an acceptable approximation can be found. In particular, in our experimental implementation we use the fourth-order Runge-Kutta method for iteratively approximating solutions of differential equations[4].

Let $rent_k = \langle l_k, t, md \rangle$ be a user rental request, $s_i$ a potential station, and $t_{exp} = t + wtime(l_k, l(s_i))$ the expected arrival time of a user $u_k$ at station $s_i$. Then the system (13)-(15) can be used to calculate the probability that $u_k$ would find an available bike (or slot) at station $s_i$ when she arrives to it at time $t_{exp}$. These probabilities are denoted as $bikeProb(s_i, rent_k)$ and $slotProb(s_i, return_k)$, respectively.

We define $\pi^i(0)$, the current probability distribution of (available) bikes locked at station $s_i$ and the current time $t$, by:

$$\pi_j^i(0) = \begin{cases} 1, & \text{if } j = bikes(s_i, t) + changes(s_i, t, t_{exp}) \\ 0, & \text{otherwise} \end{cases} \tag{16}$$

Note that similar to the *DistanceExpectedResources* strategy defined before, we include changes in the number of available bikes at the station that are expected to take place between $t$ and $t_{exp}$. In particular, we approach such changes as if they had been "blocked" already at the moment of the user request ($t$).

Using $\pi^i(0)$ as described above, setting $\lambda = returnDemand(s_i, t, t_{exp})$ and $\mu = rentDemand(s_i, t, t_{exp})$, and solving the system in the sense of (13)-(15), we can approximate a solution for the probability distribution of available bikes at the time when the user arrives at the station: $\pi^i(t_{exp})$. And $\pi^i(t_{exp})$ can be used to calculate the probability that user $u_k$ will find a bike when she arrives at station $s_i$ (at expected arrival time $t_{exp}$):

$$bikeProb(s_i, rent_k) = \sum_{j=1+committedBikes(s_i,t_{exp})}^{c(s_i)} \pi_j^i(t_{exp}) \qquad (17)$$

---

[4] Given an initial problem defined by $\frac{dy}{dt} = f(t, y)$, and $y(t_0) = y_0$, solutions for $y(t_0 + \Delta t)$ are approximated by iteratively calculating $y(t_i + h)$ based on $y(t_i)$ and $t_i$. Taking small steps $h < \Delta t$, this process is repeated $i$ steps, until $t_i + h$ is approximately $t_0 + \Delta t$. In this case, $y(t_i + h)$ approximates $y(t_0 + \Delta t)$.

As defined before, $comittedBikes(s_i, t_{exp})$ represents the maximum number of bikes that are "committed", e.g., that should be "blocked" for other users already known to arrive at station $s_i$. Thus, $bikeProb(s_i, rent_k)$ is not just the probability that a given user $u_k$ will find an available bike at a station when she arrives, but the probability of having enough bikes for the user and all other users who have been recommended to rent a bike at this station and are expected to arrive after user $u_k$. For instance, if there are a maximum of two "committed" bikes, $bikeProb(s_i, rent_k)$ would be the probability that the number of available bikes at the expected arrival time $t_{exp}$ is three or more.

In a similar way, in the case of a return request $return_k$, we need to estimate the probability that, at the expected arrival time $t_{exp}$, there is at least one slot available, plus the maximum required number of "committed" slots for future expected users. Considering the capacity of a station ($c(s_i)$), the probability of having at least $1 + committedSlots(s_i, t_{exp})$ slots is equivalent to having at most $c(s_i) - (1 + committedSlots(s_i, t_{exp}))$ locked bikes. Thus, we calculate the return probability by:

$$slotProb(s_i, return_k) = \sum_{j=0}^{c(s_i)-(1+committedSlots(s_i,t_{exp}))} \pi_j^i(t_{exp}) \qquad (18)$$

4.3.3. Expected Cost Recommendation

We combine distance and bike or slot probabilities to define the (local) rental or return cost of a station $s_i$ for a user $u_k$:

$$localRentCost(s_i, rent_k, RentFailCost) = \\ wtime(l_k, l(s_i)) + (1 - bikeProb(s_i, rent_k)) \cdot RentFailCost \qquad (19)$$

$$localReturnCost(s_i, return_k, ReturnFailCost) = \\ btime(l_k, l(s_i)) + slotProb(s_i, return_k) \cdot wtime(l(s_i), d_k) \\ + (1 - slotProb(s_i, return_k)) \cdot ReturnFailCost \qquad (20)$$

*RentFailCost* and *ReturnFailCost* are parameters that represent the associated cost of getting to a station and finding no available bike or slot, respectively. In order to avoid distortions, in case $wtime(l(s_i), d_k) > ReturnFailCost$ we set $localReturnCost(s_i, return_k, ReturnFailCost) = btime(l_k, l(s_i)) + wtime(l(s_i), d_k)$. It should also be noted that in the return cost we not only take into account the walking time from the station to the final destination, but also the cycling time to the selected station.

Using the cost functions, the definition of the *ExpectedCost* recommendation strategy simply orders stations by increasing cost:

$$RentStation(rent_k) = (s_i)_{i=1}^m, \text{ where} \qquad (21)$$
$$\forall s_i: wdist(l_k, l(s_i)) \leq md \land$$
$$\forall s_i, s_j, i < j: localRentCost(s_i, rent_k, RentFailCost)$$
$$\leq localRentCost(s_j, rent_k, RentFailCost)$$

$$ReturnStation(return_k) = (s_i)_{i=1}^m, \text{ where} \qquad (22)$$
$$\forall s_i, s_j, i < j: localReturnCost(s_i, return_k, ReturnFailCost)$$
$$\leq localReturnCost(s_j, return_k, ReturnFailCost)$$

## 5. Evaluation of User-Centred Recommendation

We used the Bike3S simulation tool [37] for validating the proposed strategies and compare their performance to the standard strategies. Bike3S is aimed to evaluate different rebalancing strategies. The simulated infrastructure includes the location, capacity, and available bikes for each station. During a simulation, users are generated and interact with the infrastructure to rent or return bikes. They may ask a service (strategy) for a recommendation to choose their target station. Several user models can be defined and used in the same simulation run. Users' appearance time, initial location and destination are loaded into the simulator. A tool facilitates the creation of this information randomly. Different strategies can be implemented and easily integrated into Bike3S.

We replicated the operation of the BiciMAD public bike sharing system in Madrid (Spain). The system covers an area of about 5x5 km of central Madrid and is continuously growing. Currently it counts on about 200 stations and 2500 bicycles. The capacity of the stations is between 12 and 30 (most stations have 24 slots).

### 5.1. Simulation Experiment Setup

There are publicly available data of the usage of the BiciMAD system, in particular:

- Data of the trips: including, for each trip, time of taking a bike, origin station, destination station, travel time and the approximate route. However, in order to anonymize the data, only the day and hour of the pick-up time of each trip are given (without minutes). Each trip includes a user type, with possible values representing regular or occasional users, BiciMAD staff or unidentifiable users.
- Situation of the stations: including the number of available bikes and slots and whether or not a station was active.

In order to replicate a real-world scenario, we extracted the user data for a 24-hour period (in particular, from 7:00 on the 20th of July 2018 to 7:00 on the 21st of July 2018). That period comprises 12296 real user trips, whose data was used to generate the artificial users of our simulator. This data, however, only includes the stations where a user took or returned a bike but not her origin or destination location. Therefore, and in order to reflect the real situation as accurately as possible, we generate random origin and destination locations in the area spanned by a circle of 300 meters around the rental station and the return station, respectively. Finally, the instant of creation of the simulated user is generated randomly within the hour of her appearance in the real-world data (with a uniform distribution), since the exact minute and second of appearance is unknown.

For specifying the initial station configuration in the simulated scenario, we used the official (real) data at the initial time (20th of July 2018 at 7:00). There have been 169 active stations with a total capacity of 4086 slots and 1792 bikes plugged in. However, since we want to analyse the performance of the different recommendation strategies in rather critical situations (where it is more likely to have empty and full stations), we reduced the station capacities and the number of plugged bikes at each station to approximately half of the original values. That is, in our experiments we used 169 stations with a total capacity of 1995 slots and 896 bikes.

Even though the Bike3S tool can simulate the movements of people walking or cycling on the real network of roads and walk paths (using OpenStreetMap data), in our simulation we used straight-line movements on the geographical map. The impact of this choice on performance comparison of the analysed strategies is marginal, but it allowed us to reduce the simulation time considerably. We consider a default walking and cycling velocity of 1.4 m/s and 4 m/s, respectively [38,39]. However, since we use straight-line movements and in order to adjust to more realistic values, we apply a velocity factor of 0.614. This factor has been established based on comparing real and straight-line travel times between a set of origin/destination locations in Madrid (using OpenStreetMap data). Thus, the actually applied velocities are 0.8596 m/s for walking and 2.456 m/s for cycling.

This velocity is used to simulate the movements of users. It is also used in the recommendation strategies to estimate the expected arrival times of users.

Using the specified setup, we carried out experiments to compare the performance of the five previously defined recommendation strategies:

- *ShortestDistance (SD)*
- *InformedShortestDistance (ISD)*
- *DistanceResources (DR)*
- *DistanceExpectedResources (DER)*
- *ExpectedCost (EC(x,y))*, where $x = RentFailCost$ and $y = ReturnFailCost$

The user behaviour for renting a bike in the simulations is as follows:

1. A user $u_k$ appears at a geographical location $l_k$ at time $t$ and asks the recommendation system for a rental recommendation (with request $rent_k = \langle l_k, t, md \rangle$). The maximum acceptable distance in the experiments is set to $md = 600$ meters.
2. The recommendation system applies its strategy and returns a ranking of possible stations.
3. Given the ranking, the user filters out all stations she has already tried. If no stations are left, the user will abandon the system without renting a bike. Otherwise, she walks towards the first station in the list of recommendations in order to rent a bike.
4. In case the user gets to a station and there are no available bikes, she repeats the whole process until she either abandons or finally finds a bike. In this case, the value of the maximum acceptable distance ($md$) is reduced by the distance the user has walked already.

With this behaviour, a user will effectively abandon the system if she does not find any bikes within $md = 600$ meters.

For finding a return station, the process is as follows:

1. In the moment a user $u_k$ has rented a bike at a station $s_i$, she issues a return request $return_k = \langle l_k, t, l_d \rangle$, where $l_k$ and $t$ are the current position and time, and $l_d$ is her final destination location.
2. The recommendation system returns a ranking of stations for returning the bike.
3. Given the ranking, the user filters out stations she has already tried and selects the first remaining station for returning the bike.
4. In case the user gets to a station and there are no available slots, she repeats the whole process until she finally finds a station to leave the bike (there is no possibility to abandon).

*5.2. Simulation Results*

The main aim of the recommendation strategies is to allow an efficient usage of a BSS in terms of the time the users spend to go from their origin location to their destination, as well as the unsuccessful rental and return attempts.

Table 3 compares the performance of the different user-centred recommendation strategies. The measurements we present are the following:

- #a: number of users who dropped out (abandoned) the system and percentage with regards to users that finished
- #fh: number of failed user rental attempts and percentage over all user rental attempts
- #fr: number of failed user return attempts and percentage over all user return attempts
- *tt*: average total time of users in the system; it is based on the time a user requires to go from her origin to a bike rental station, to cycle from there to a station to return the bike, and finally to walk to her final destination. The value is averaged over all users that were able to rent a bike (i.e., who did not abandon).

- *AET*: average station empty time; this is the time a station has been empty (without available bikes), and thus, would potentially have been denying a service. The value is the average over all 169 stations for the whole simulation period.

**Table 3.** Experiments results for Madrid. Bold numbers indicate the best obtained result for each metric.

| Strategy | #a/% | #fh/% | #fr/% | tt (min) | AET (min) |
|---|---|---|---|---|---|
| OPTIMUM | 0 | 0 | 0 | 19.21 | |
| SD | 1573/12.85 | 1834/14.67 | 5148/32.6 | 22.47 | 378.4 |
| ISD | 895/7.3 | 415/3.53 | 1965/14.7 | 21.86 | 398.1 |
| DR | 461/3.76 | 126/1.06 | 225/1.9 | 21.54 | 201.9 |
| DER | 243/1.98 | **0/0** | **0/0** | 21.62 | 189.7 |
| EC(1000/2000) | 428/3.49 | 72/0.6 | 269/2.2 | **21.12** | 362.7 |
| EC(3000/2000) | 367/2.99 | 13/0.11 | 183/1.5 | 21.24 | 309.0 |
| EC(70000/2000) | 240/1.96 | 3/0.02 | 260/2.1 | 21.6 | 197.5 |
| EC($10^6$/$10^6$) | **148/1.21** | 1/0.01 | **0/0** | 23.21 | **75.7** |

The first line of Table 3 specifies a hypothetical optimal situation, where every user can obtain a bike at the station closest to her origin and return it at the station closest to her destination (i.e., the station capacities for rental and return are assumed to be illimited). In this case, the average travel time would be 19.21 minutes. This travel time can obviously not be obtained in a real case.

Analysing the standard strategies (the first 4 strategies in Table 3), there are different conclusions.

*SD* has the worst behaviour. Users here, go to the closest station and many users will abandon because there are no bikes available. Furthermore, the travel time is quite high, because after an unsuccessful rental or return attempt, users may try at other stations and, thus, increment the time they spend in the system. In comparison, *ISD* has a much better performance, because the probability to find an available bike or slot is much higher if a user only goes to the stations that have available resources at the moment of the recommendation request.

*DR* combines distance and availability of resources. In this way, it obtains a better dropout rate and at the same time a shorter travel time than *SD* and *ISD*. It should also be noted that here, users would slightly prefer to rent/return bikes where there are more bikes or slots (within closed stations). That means, here, users implicitly improve the distribution of the bikes (and slots) among the stations. This can be seen in the reduction of the average empty time of stations.

The *DistanceExpectedResources* strategy (*DER*) clearly outperforms the standard strategies. Compared to *DR*, *DER* not only includes the bikes/slots in the analysis of appropriate stations that are available at the precise moment when a user issues a recommendation request, but also the expected changes up to the expected arrival of the user. In this case, the dropout rate can be reduced from 3.76% to 1.98% whereas the average travel time is nearly the same (21.62 versus 21.54 minutes). It should be noted that with *DER*, the failed rental and return attempts are both 0. This is because with the specified experimental setting, the estimation of available bikes/slots is exact because (i) users always go to the recommended stations, and (ii) the estimated travel time of users is the same as their actual travel time. In a real scenario, the benefit of *DER* wrt. *DR* should be somewhat smaller.

The last four lines in Table 3 correspond to different instances of the *ExpectedCost* strategy (*EC*) that prioritises stations with a lower cost, calculated based on travel time and the (demand dependent) probability of finding a bike/slot at the expected time of arrival of the user. Its performance varies for different values of *RentFailCost* and *ReturnFailCost*. In general, higher penalizations produce a lower abandon rate but increase the travel time. With $RentFailCost = ReturnFailCost = 1000000$, for example, the abandon rate is reduced to 1.21% but the travel time increases to 23.21 minutes, which is considerable. A rather high value for *RentFailCost* and low value for *ReturnFailCost* obtains a good compromise. In this sense, for *RentFailCost* = 70000 and *ReturnFailCost* = 2000 both, dropout rate and travel time, can be slightly reduced with regard to *DER*. With rather low penalizations, the abandon rate is higher but the travel time can be reduced, as with *EC(1000,2000)*.

In Figure 2, Figure 3 and Figure 4 we analyse the behaviour of *EC* with varying values for *RentFailCost* and *ReturnFailCost*.

As shown in Figure 2, the abandon rate decreases with increasing *ReturnFailCost* and the travel time up to a value of about 2000 decreases and then increases. At higher values both, travel time and abandon rate become almost stable. The lowest travel time is obtained with a value of about 2000. Exactly the same behaviour is observed for other values of *RentFailCost*. We also observed in our experiments that, as the value of *ReturnFailCost* increases, the number of failed returns (#fr) falls rapidly towards about 0 at the value of *ReturnFailCost* = 4000 (for the sake of clarity, this is not shown in the figure).

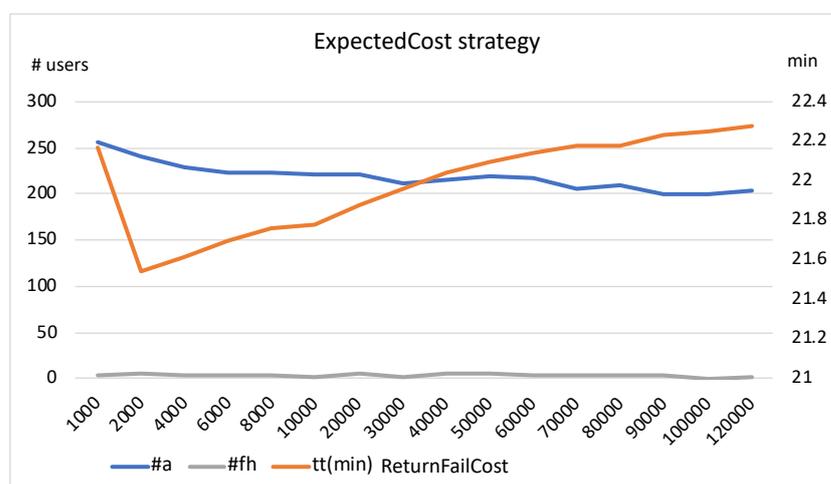

**Figure 2.** *ExpectedCost* strategy with fixed *RentFailCost* = 50000 and varying *ReturnFailCost*..

In Figure 3 we fixed *ReturnFailCost* = 2000 (the value that obtains generally the lowest travel time for different values of *RentFailCost*) and we increase *RentFailCost*. As it can be seen, higher values of *RentFailCost* obtain lower abandon rates and less failed rental attempts but lead to increased travel times. It seems that the values remain almost stable at a certain point.

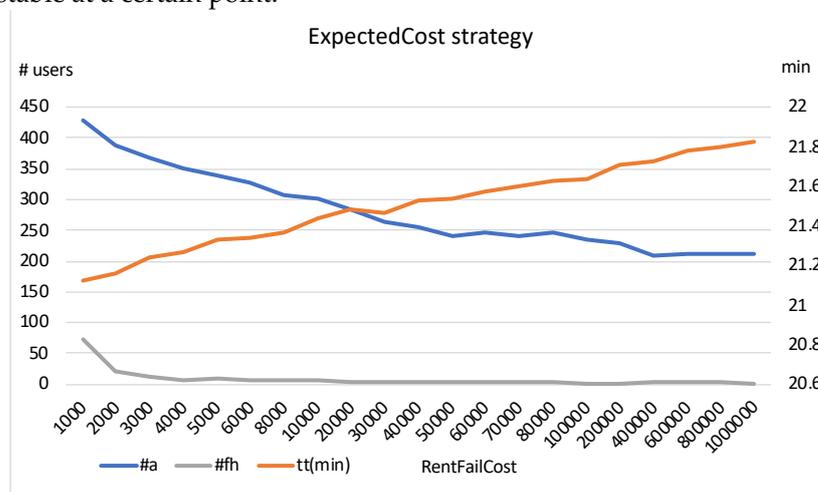

**Figure 3.** *ExpectedCost* strategy with fixed *RentFailCost* and fixed *ReturnFailCost* = 2000.

If we increment *RentFailCost* and *ReturnFailCost* at the same time as in Figure 4, the abandon rate reduces continuously and considerably, but at the cost of an increase in travel time. The failed rental attempts tend towards 0.

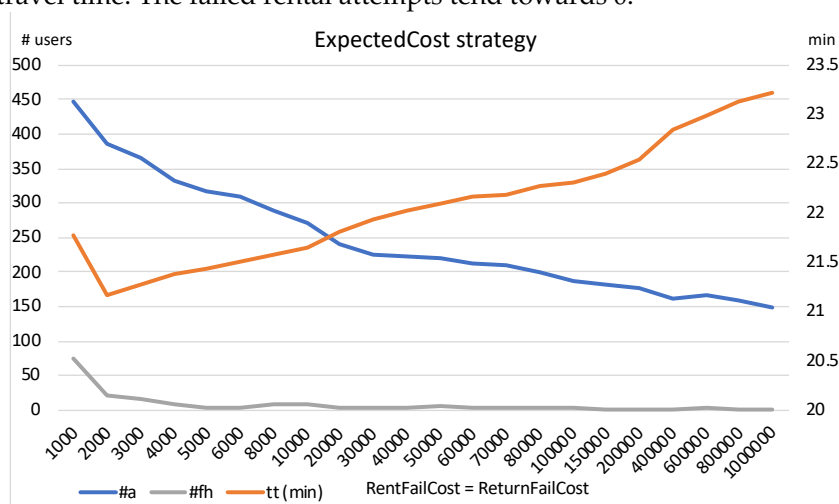

**Figure 4.** *ExpectedCost* strategy with increasing equal values of *RentFailCost* and fixed *ReturnFailCost*.

Comparing the *ExpectedCost* strategy with *DistanceExpectedResources*, the former can be parametrized in order to prioritize abandon rate or travel time. However, it is interesting to observe that for the same ratio of travel time, both methods obtain almost the same dropout rate and vice versa. In other words, the rather simple *DER* strategy performs as well as the much more sophisticated *EC* strategy. In our opinion, the explanation of this fact is that the *DER* strategy not only proposes an appropriate station to rent or return a bike in terms of the utility for an individual user. In addition, this method promotes that users take or return bikes at stations with more bikes/slots. In this way, the strategy implicitly improves the unbalancing problem. It helps to obtain a better distribution of bikes (and slots) and thus, to reduce the abandon rate and the travel time of potential future users.

## 6. Recommendation Based on Local and Global Utility

As argued before, the *DER* strategy with a rather simple calculation model has a quite acceptable performance, because it combines the local utility of a recommendation for a user with a certain improvement of the global situation for the future. In this section we use this idea to extend the *ExpectedCost* strategy in the same sense. The idea is not only to account for the cost of a station for a certain user, but also for the changes of the costs of potential future users, if that decided to take (or return) a bike at given station. In particular, taking into account the future demand of bikes or slots at a station, we estimate how taking or returning a bike will increase or decrease the number of expected unsuccessful rental and return attempts of future users at a station within a certain time frame.

### 6.1. Calculating the Future Impact of Rentals and Returns

To estimate the impact that a bike rental or return has globally on system performance, we analyse how such an event would change the number of expected unsuccessful rental and return attempts of future users. That is, the impact can be measured through the difference in expected unsuccessful events during some time interval $[t_a, t_b]$, if a bike is taken (or returned) at the beginning of this time interval. Moreover, given a station $s_i$ and the expected rental rate during the interval $[t_a, t_b]$ (obtained by analysing historical demands, for example) the number of expected unsuccessful rental attempts is the number of all rental attempts multiplied by the probability that such attempts fail. We formalize this idea in the sequel.

Modelling a station $s_i$ as a queue, the average probability of the station being in state 0, that is having no available bike, during a time interval $[t_a, t_b]$ with $t_b > t_a$ is given by:

$$\bar{\pi}_0^i(t_a \to t_b) = \frac{1}{t_b - t_a} \int_{t_a}^{t_b} \pi_0^i(w) \, dw \qquad (23)$$

Based on the system of differential equations (13)-(15) and given the initial probability distribution $\pi^i(t_a)$ and a step-size $h$ with $(t_b - t_a) > h > 0$, we can use the fourth-order Runge-Kutta method (RK4) to estimate $\bar{\pi}_0^i(t_a \to t_b)$. Let $\left(\pi_0^i(t_a), \pi_0^i(t_a + h), \pi_0^i(t_a + 2h), \ldots, \pi_0^i(t_a + kh)\right)$, where $t_a + kh = t_b$, denote the sequence of iteratively calculated values for $\pi_0^i(t)$ from $t = t_a$ to $t_b$ at steps $h$ with RK4. Now we can use a Riemann sum to approximate $\bar{\pi}_0^i(t_a \to t_b)$:

$$\bar{\pi}_0^i(t_a \to t_b) = \frac{1}{t_b - t_a} \sum_{j=0}^{k-1} h \frac{\pi_0^i(t_a + jh) + \pi_0^i(t_a + (j+1)h)}{2} \qquad (24)$$

Finally, given the rental appearance rate for a station $s_i$ in the interval $[t_a, t_b]$, $rentDemand(s_i, t_a, t_b)$, the expected number of failed rental attempts in the interval is estimated by:

$$ehf(s_i, t_a, t_b) = \bar{\pi}_0^i(t_a \to t_b) \cdot rentDemand(s_i, t_a, t_b) \qquad (25)$$

In a similar way we can calculate the expected failed return attempts for a station $s_i$ in an interval $[t_a, t_b]$:

$$erf(s_i, t_a, t_b) = \bar{\pi}_{c(s_i)}^i(t_a \to t_b) \cdot returnDemand(s_i, t_a, t_b) \qquad (26)$$

where $c(s_i)$ is the capacity of station $s_i$.

If a user $u_k$ issues a rental request $rent_k = \langle l_k, t, md \rangle$, we use $ehf$ and $erf$ to estimate the future impact that a potential rental would have at a station $s_i$. The idea is to calculate first the probability distribution of available bikes, when the user arrives at the station. Afterwards, we analyse the expected rental and return failures that would arise after the arrival and during a given timeframe ($tf$) if the user rented out a bike and if she did not do so. The difference between these values can be used as a measure for estimating the future impact of renting a bike at a station. Formally, we first calculate the probability distribution of bike availability at the expected arrival time of user $u_k$: $\pi^i(t_{exp})$, with $t_{exp} = t + wtime(l_k, l(s_i))$ as explained in Section 4. Using $\pi^i(t_{exp})$ as initial distribution and applying (24), (25) and (26), we can calculate $ehf(s_i, t_{exp}, t_{exp} + tf)$ and $erf(s_i, t_{exp}, t_{exp} + tf)$. Then, we transform from $\pi^i(t_{exp})$ to a distribution $\pi^{i,rent}(t_{exp})$, where one bike is "taken" away. That means:

$$\pi_0^{i,rent}(t_{exp}) = \pi_0^i(t_{exp}) + \pi_1^i(t_{exp}) \qquad (27)$$
$$\pi_k^{i,rent}(t_{exp}) = \pi_{k+1}^i(t_{exp}), \quad \text{for } c(s_i) > k > 0 \qquad (28)$$
$$\pi_k^{i,rent}(t_{exp}) = 0, \quad \text{for } k = c(s_i) \qquad (29)$$

Now, taking $\pi^{i,rent}(t_{exp})$ as initial distribution, we calculate $ehf^{rent}(s_i, t_{exp}, t_{exp} + tf)$ and $erf^{rent}(s_i, t_{exp}, t_{exp} + tf)$. Finally, we can obtain the rental failure impact ($ehfIm^{rent}$) and return failure impact ($erfIm^{rent}$) of the potential rental:

$$ehfIm^{rent}(u_k, s_i, tf) = ehf^{rent}(s_i, t_{exp}, t_{exp} + tf) - ehf(s_i, t_{exp}, t_{exp} + tf) \qquad (30)$$

$$erfIm^{rent}(u_k, s_i, tf) = erf^{rent}(s_i, t_{exp}, t_{exp} + tf) - erf(s_i, t_{exp}, t_{exp} + tf) \qquad (31)$$

It should be noted that $ehf^{rent}(s_i, t_{exp}, t_{exp} + tf) \geq ehf(s_i, t_{exp}, t_{exp} + tf)$ because the expected rental failure rate is necessarily higher if a bike is taken from a station, thus $ehfIm^{rent}(u_k, s_i, tf)$ would be positive. On the other hand, $erfIm^{rent}(u_k, s_i, tf) \leq 0$, since the number of available slots increases if a bike is taken from the station. Thus, the values of $ehfIm^{rent}(u_k, s_i, tf)$ and $erfIm^{rent}(u_k, s_i, tf)$ represent the expectation of failed rental and return attempts, that are caused by user $u_k$ when she takes a bike at station $s_i$.

In a similar way, but transforming $\pi^i(t_{exp})$ to a distribution $\pi^{i,return}(t_{exp})$, where one bike is "added" allows us to calculate the expected failed rental and return attempts, that are caused if user $u_k$ returns a bike at station $s_i$:

$$ehfIm^{return}(u_k, s_i, tf) = ehf^{return}(s_i, t_{exp}, t_{exp} + tf) - ehf(s_i, t_{exp}, t_{exp} + tf) \quad (32)$$

$$erfIm^{return}(u_k, s_i, tf) = erf^{return}(s_i, t_{exp}, t_{exp} + tf) - erf(s_i, t_{exp}, t_{exp} + tf) \quad (33)$$

*6.2. Recommendation Based on Expected Cost and Future Impact*

We calculate the global cost of renting at a station by combing its local cost with the impact on the overall system in the next given time frame:

$$\begin{aligned}globalRentCost(s_i, rent_k, tf, RentFailCost, FRentFC, FReturnFC) = \\ localRentCost(s_i, rent_k, RentFailCost) + f \cdot bikeProb(s_i, rent_k) \cdot \\ (ehfIm^{rent}(u_k, s_i, tf) \cdot rentCost(s_i, l_k, tf, FRentFC) + \\ erfIm^{rent}(u_k, s_i, tf) \cdot returnCost(s_i, l_k, tf, FReturnFC))\end{aligned} \quad (34)$$

$$\begin{aligned}globalReturnCost(s_i, return_k, tf, ReturnFailCost, FRentFC, FReturnFC) = \\ localReturnCost(s_i, return_k, ReturnFailCost) + f \cdot slotProb(s_i, return_k) \cdot \\ (ehfIm^{return}(u_k, s_i, tf) \cdot rentCost(s_i, l_k, tf, FRentFC) + \\ erfIm^{return}(u_k, s_i, tf) \cdot returnCost(s_i, l_k, tf, FReturnFC))\end{aligned} \quad (35)$$

$f$ is a parameter to give more or less importance to the local cost versus the global impact. $rentCost$ and $returnCost$ represent the cost associated to the number of extra future rental and return failures. This cost should depend on whether a potential user who arrives at a station and fails to rent or to return a bike has some alternative to rent or return a bike at a station close by. With this idea, we define:

$$rentCost(s_i, l_k, tf, FRentFC) = \begin{cases} \min_{s_j \epsilon S \,\wedge wdist(s_i,s_j) \leq MD} localRentCost(s_j, <l(s_i), t', MD>, FRentFC) \\ FRentFC, \quad if \neg \exists s_j \epsilon S : wdist(s_i, s_j) \leq MD \end{cases} \quad (36)$$

$$returnCost(s_i, l_k, tf, FReturnFC) = \begin{cases} \min_{s_j \epsilon S \,\wedge bdist(s_i,s_j) \leq MD} localReturnCost(s_j, <l(s_i), t'', MD>, FReturnFC) \\ FReturnFC, \quad if \neg \exists s_j \epsilon S : bdist(s_i, s_j) \leq MD \end{cases} \quad (37)$$

where $t' = t + wtime(l_k, l(s_i)) + tf/2$ and $t'' = t + btime(l_k, l(s_i)) + tf/2$. This means, $rentCost$ is the lowest local cost of stations in the neighborhood of station $s_i$ (within a distance of $MD$ meters) and taking the parameter $FRentFC$ as the penalization cost. In the worst case, e.g when no station exists within a distance of $MD$ meters from $s_i$, the value of $rentCost$ will be $FRentFC$. In the same way, $returnCost$ will be at most $FReturnFC$. Note that the local costs are approximated for users who would arrive at the station $s_i$ at half of the timeframe after the arrival of the user $u_k$.

Based on the global costs, we define the recommendation strategy *ExpectedCostFutureImpact*:

$$RentStation(rent_k) = (s_i)_{i=1}^{m}, \text{ where} \quad (38)$$
$$\forall s_i : wdist(l_k, l(s_i)) \leq md \land$$
$$\forall s_i, s_j, i < j:$$
$$globalRentCost(s_i, rent_k, tf, RentFailCost, FRentFC, FReturnFC) \leq$$
$$globalRentCost(s_j, rent_k, tf, RentFailCost, FRentFC, FReturnFC)$$

$$ReturnStation(return_k) = (s_i)_{i=1}^{m}, \text{ where} \quad (39)$$
$$\forall s_i, s_j, i < j:$$
$$globalReturnCost(s_i, return_k, tf, ReturnFailCost, FRentFC, FReturnFC) \leq$$
$$globalReturnCost(s_j, return_k, tf, ReturnFailCost, FRentFC, FReturnFC)$$

In summary, the parameters of this strategy are:

- $md$ - the maximum distance to rental stations
- $RentFailCost$ and $ReturnFailCost$ - the penalization costs applied if a user is unsuccessful when trying to rent or return a bike
- $tf$ - the timeframe for predicting the future impact of a rental or return action
- $FRentFC$ and $FReturnFC$ - the penalization costs applied when estimating the costs of local alternative stations for users that would not find a bike or slot at the station $s_i$ in the future
- $MD$ - the maximum distance to consider alternative stations for users that would not find a bike or slot at the station $s_i$
- $f$ - the factor applied to the global impact in the cost estimation

## 7. Evaluation of *ExpectedCostFutureImpact* Recommendation

We used the same settings as in Section 5 to evaluate the *ExpectedCostFutureImpact* strategy. In the experiments, we fixed the following parameters: the values of $md = 600$ meters as a reasonable distance a user might be willing to walk to get a bike and of $MD = 500$ meters.

In the first set of experiments, we analysed this approach with different penalization costs, and fixing the timeframe $tf$ to 1 hour and the factor $f$ =1. Some results are presented in Table 4.

Analysing the results presented in the table, we observe that in general higher penalization costs lead to lower dropout rates but longer travel times, whereas lower costs have the opposite effect: more abandons and shorter travel times. A good combination of both metrics can be obtained with values close to the following: $RentFailCost$ =3000, $ReturnFailCost$ =2500, $FutRentFailCost$ =3000 and $FutReturnFailCost$ =1000. Especially the abandon rate can be reduced considerably if a recommendation not only considers the local cost for a user (to rent or return a bike), but also the impact on potential users in the future. For example, the strategy with the penalization schema 3000/2500/3000/1000, as compared to *EC(70000/2000)*, reduces the number of abandons from 240 to 15 by maintaining the same travel time (21.6 minutes). In Section 5, the *EC(70000/2000)* strategy showed a good behaviour in terms of abandon rate and low travel time.

**Table 4.** Experiments results for Madrid. Bold numbers indicate the best obtained result for each metric.

| *RentFailCost/Return-FailCost/FutRentFailCost/FutReturnFailCost* | #a | #fh | #fr | tt (min) | AET (min) |
|---|---|---|---|---|---|
| 1000/2000/1000/2000 | 193 | 71 | 87 | 21.04 | 223.3 |
| **3000/2500/3000/1000** | 15 | 2 | 20 | 21.63 | 128.0 |
| 3000/2000/3000/2000 | 29 | 3 | 83 | 21.70 | 132.5 |
| 70000/2000/70000/2000 | 27 | 0 | 734 | 29.25 | 24.22 |
| 1000000/1000000/1000000/1000000 | 22 | 0 | 209 | 40.61 | 18.13 |
| 100000/100000/100000/100000 | 10 | 0 | 162 | 28.99 | 11.67 |
| 50000/50000/50000/50000 | 14 | 0 | 124 | 27.08 | 13.11 |
| 10000/10000/10000/10000 | 20 | 1 | 15 | 23.27 | 40.79 |
| 5000/5000/5000/5000 | 22 | 1 | 1 | 22.37 | 83.30 |
| 3000/3000/3000/3000 | 35 | 1 | 6 | 21.80 | 127.9 |

Using the penalization cost pattern 3000/2500/3000/1000, and fixing the factor $f$=1, in Figure 5 we analyse the influence of the prediction timeframe in the results.

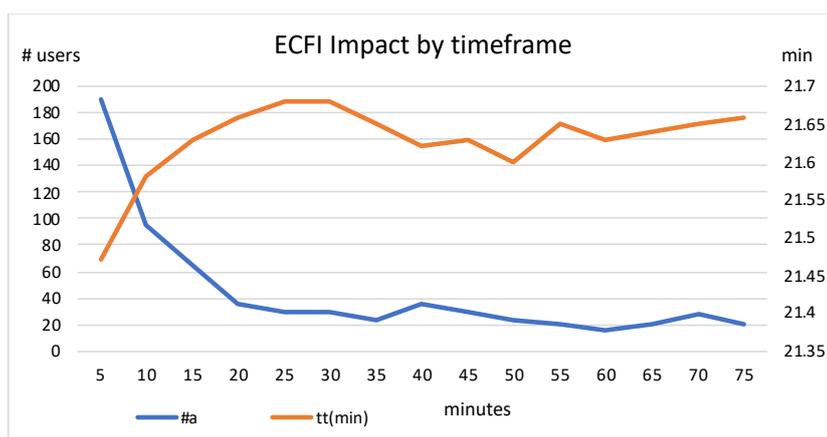

**Figure 5.** Impact of different prediction timeframes in the behaviour of *ExpectedCostFutureImpact* (with *RentFailCost*=3000, *ReturnFailCost*=2500, *FutRentFailCost*=3000, *FutReturnFailCost*=1000, and $f$=1).

Figure 5 indicates that, as expected, short prediction timeframes present higher abandon rates but faster travel times. It is more likely that future users will not find an available bike or slot, but users are sent to "closer" stations since the considered future impact will be lower. At some point, the values for dropout rate and travel time tend to stabilize and even increase slightly. Good results are obtained with a prediction timeframe of about 20 minutes.

The best way to prioritize between average travel time and abandon rate is via the factor $f$. In Figure 6 we show different results obtained when varying $f$. Low values for $f$ decrease travel time but lead to more abandons. A good compromise is obtained with $f$ around 1.

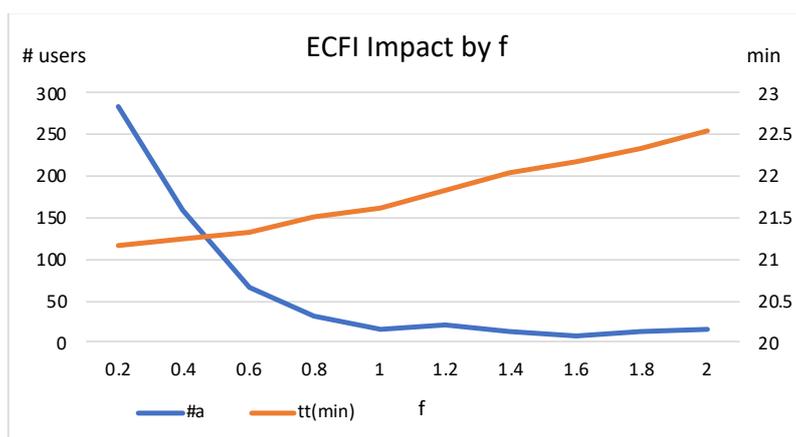

**Figure 6.** Impact of different factors $f$ with *ExpectedCostFutureImpact* (with *RentFailCost*=3000, *ReturnFailCost*=2500, *FutRentFailCost*=3000, *FutReturnFailCost*=1000, and $tf$=60 minutes.

As we mentioned in Section 5, the experiments in this paper are based on real data, but we reduced the station capacities and the overall number of bikes in the system to about half of the original values. The reason is that such a situation is more challenging and the differences in the recommendation strategies are more appreciable. In Table 5, however, we present and compare the standard strategies with the best variations of the *ExpectedCost* and the *ExpectedCostFutureImpact* strategies in a scenario with the original (real) station capacities and bike numbers. As expected, all strategies significantly improved their performance with real station capacities. As shown, also in this situation, the novel strategy *ExpectedCostFutureImpact* proposed in this paper performs better than the other strategies, in both, abandon rate and travel time.

**Table 5.** Comparison of strategies with real station capacities and number of bikes.

| Strategy | #a | #fh | #fr | tt (min) | AET (min) |
|---|---|---|---|---|---|
| OPTIMUM | 0 | 0 | 0 | 19.21 | |
| SD | 262 | 697 | 1613 | 20.22 | 143.75 |
| ISD | 115 | 90 | 437 | 19.85 | 145.75 |
| AR | 24 | 0 | 17 | 27.99 | 3.98 |
| DR | 38 | 7 | 18 | 20.01 | 37.55 |
| DER | 18 | 0 | 0 | 20.01 | 32.87 |
| EC(3000/2000) | 41 | 0 | 1 | 19.52 | 87.89 |
| ECFI(*RentFailCost*=3000, *ReturnFailCost*=2500, *FutRentFailCost*=3000, *FutReturnFailCost*=1000, $tf$=60 min and $f$=1) | 6 | 0 | 0 | 19.55 | 31.46 |

## 8. Conclusion

Bike sharing systems are becoming an integral part of intelligent transportation infrastructures in Smart Cities. Station-based BSS have the advantage of being more resilient with regards to misuse and vandalization, and also account for seamless charging if the BSS fleet contains electric bicycles. A typical problem in station-based BSS is the possibility that some stations run out of available resources due to high and unbalanced demands at peak times. However, if no bikes are available near their location or if finding an available parking slot at the destination is difficult, users may drop out of the BSS and use other, more contaminating means of transportation.

In this paper we address the balancing problem in BSS by developing recommendation strategies that help users select a station to rent or return a bike considering the distance/time to that station as well as the likelihood that they will find a bike/slot when they actually arrive at the station.

Our contribution is two-fold. In the first part of the paper, we presented and analysed station selection (or recommendation) strategies that are user-centred, that is, they try to find the best station considering only the utility or expected cost of a specific user. We presented the *DistanceExpectedResources* strategy, which assumes that recent recommendations are being followed by users and thus, can better estimate the resources that are expected to be available at the moment when a user actually arrives at a certain station. We also presented the *ExpectedCost* strategy that minimizes a user's cost, combining the distance from her origin (or destination) to the location of candidate stations, and the probability of finding an available bike (or empty slot) when she arrives. This strategy models stations as queues and uses demand data to estimate the probabilities of finding available bikes or slots. We compared the performance of the presented strategies through simulation experiments with real data from the BiciMAD BSS in Madrid. Both methods outperform baseline station selection strategies like "going to the closest" or "going to the closest station with available bikes / slots" in terms of (i) the number of users who abandon the system without renting a bike and (ii) the total time in the system.

In the second part of the paper, we proposed a station recommendation strategy that seeks an equilibrium between *local (user-centred)* and *global* utility. With regard to the latter, recommendations prioritize stations that are good for a particular user, but also imply some positive impact on the distribution of bikes and slots in the overall system and for potential future users. In particular, we defined the *ExpectedCostFutureImpact* strategy that extends the *ExpectedCost* approach by analysing also the impact that choosing a particular station will have on future rentals and returns. In the simulation experiments, this solution outperforms all other strategies, both with respect to the number of abandons and to total travel time.

In future work we aim to look into explicit (e.g., monetary) incentive mechanisms so as to persuade individually rational users to follow our recommendations while maintaining their trust in the system. These incentives are likely to be proportional in some way to the *globalRentCost*. Another interesting line for future research in this context consists in learning from experience the likelihood that specific (types of) users will follow the recommendations given. This would allow a more fine-grained adjustment of our recommendation model based on user profiles, and could also be used to implement specific user-centered incentives. Finally, as we mentioned in the introduction, BSS could be conceived as part of a sophisticated multi-modal intelligent transportation solutions. The synergic effects within such a System of Systems open up a whole range of new opportunities especially with regards to better availability of green transportation services and a reduced number of dropouts.

**Author Contributions:** Conceptualization, H.B., A.F. and S.O.; methodology, H.B., A.F. and S.O.; software, H.B.; validation, H.B., A.F. and S.O.; formal analysis, H.B., A.F. and S.O.; investigation, H.B., A.F. and S.O.; writing—original draft preparation, H.B., A.F. and S.O.; writing—review and editing, H.B., A.F. and S.O.; visualization, H.B., A.F. and S.O.; funding acquisition, H.B., A.F. and S.O. All authors have read and agreed to the published version of the manuscript.

**Funding:** This work has been partially supported by the Spanish Ministry of Science, Innovation and Universities, co-funded by EU FEDER Funds, through grant RTI2018-095390-B-C33 (MCIU/AEI/FEDER, UE).

**Conflicts of Interest:** The authors declare no conflict of interest.